%% file: acl_latex.tex
\pdfoutput=1
\documentclass[11pt]{article}

\usepackage[final]{acl}

\usepackage{times}
\usepackage{latexsym}
\usepackage{enumitem}

\usepackage[T1]{fontenc}
\usepackage[utf8]{inputenc}

\usepackage{microtype}

\usepackage{inconsolata}

\usepackage{graphicx}

\usepackage{adjustbox}
\usepackage{array}
\usepackage{booktabs}
\usepackage{makecell}
\usepackage{multirow, multicol}
\usepackage{amsmath}
\usepackage{threeparttable}

\title{Overview of the EHRSQL 2024 Shared Task on \\ Reliable Text-to-SQL Modeling on Electronic Health Records}

\author{
    Gyubok Lee\quad
    Sunjun Kweon\quad
    Seongsu Bae\quad
    Edward Choi \\
    KAIST AI\\
    \texttt{\{gyubok.lee, sean0042, seongsu, edwardchoi\}@kaist.ac.kr} \\
    }

\begin{document}
\maketitle
\begin{abstract}
Electronic Health Records (EHRs) are relational databases that store the entire medical histories of patients within hospitals. They record numerous aspects of patients' medical care, from hospital admission and diagnosis to treatment and discharge. While EHRs are vital sources of clinical data, exploring them beyond a predefined set of queries requires skills in query languages like SQL. To make information retrieval more accessible, one strategy is to build a question-answering system, possibly leveraging text-to-SQL models that can automatically translate natural language questions into corresponding SQL queries and use these queries to retrieve the answers. 
The EHRSQL 2024 shared task aims to advance and promote research in developing a question-answering system for EHRs using text-to-SQL modeling, capable of reliably providing requested answers to various healthcare professionals to improve their clinical work processes and satisfy their needs.
Among more than 100 participants who applied to the shared task, eight teams were formed and completed the entire shared task requirement and demonstrated a wide range of methods to effectively solve this task. In this paper, we describe the task of reliable text-to-SQL modeling, the dataset, and the methods and results of the participants. We hope this shared task will spur further research and insights into developing reliable question-answering systems for EHRs.
\end{abstract}

\section{Introduction}

Electronic Health Records (EHRs) store all types of medical events that occur in the hospital, including admissions, diagnoses, procedures, prescriptions, and discharges.
They replace traditional paper-based records and provide a centralized repository for patient data.
Over the years, the widespread adoption of EHRs in hospitals has been shown to improve patient care, increase efficiency, and enhance coordination among healthcare professionals~\cite{upadhyay2022qualitative, mullins2020health, uslu2021value}.
Although EHRs are a valuable source of patient data, the complexity of their data structures and the need for specialized skills, such as query languages like SQL, to extract and analyze the information, often hinder their effective utilization by healthcare professionals~\cite{wang2020text, lee2022ehrsql}.
These barriers lead to the underutilization of the full potential of EHRs in clinical practice and research.

% An alternative way to utilize data stored in EHRs, without the need for query languages, is to develop a question-answering (QA) system.
% QA systems provide a user-friendly interface that allows healthcare professionals to ask questions in natural language and receive relevant answers extracted from the EHR data.
% These models automatically convert natural language questions into their corresponding SQL queries, and then use these queries to obtain the final answer.

An alternative way to utilize data stored in EHRs is to develop a question-answering (QA) system.
QA systems provide a user-friendly interface that allows healthcare professionals to ask questions in natural language and receive answers (e.g., simple patient information retrieval or complex synthesis of patient information across multiple tables) from the EHR data without needing to know query languages or EHR database structures.
Specifically, text-to-SQL modeling is an effective approach for building such QA systems for EHRs, which are typically relational databases.
These models automatically convert natural language questions into their corresponding SQL queries and then execute these queries on the database to obtain the final answer.
With the impressive advances in large language models (LLMs), various high-performance text-to-SQL models have been introduced, which are accomplished through model fine-tuning~\cite{scholak2021picard} or LLM prompting with demonstrations~\cite{pourreza2024din, gao2023text, chang2023prompt}.
If deployed with reliable performance, the adoption of these models could significantly benefit healthcare professionals by allowing them to explore patient data more freely from the EHRs through natural language interactions.

\begin{figure*}[t!]
\centering
\includegraphics[width=\textwidth]{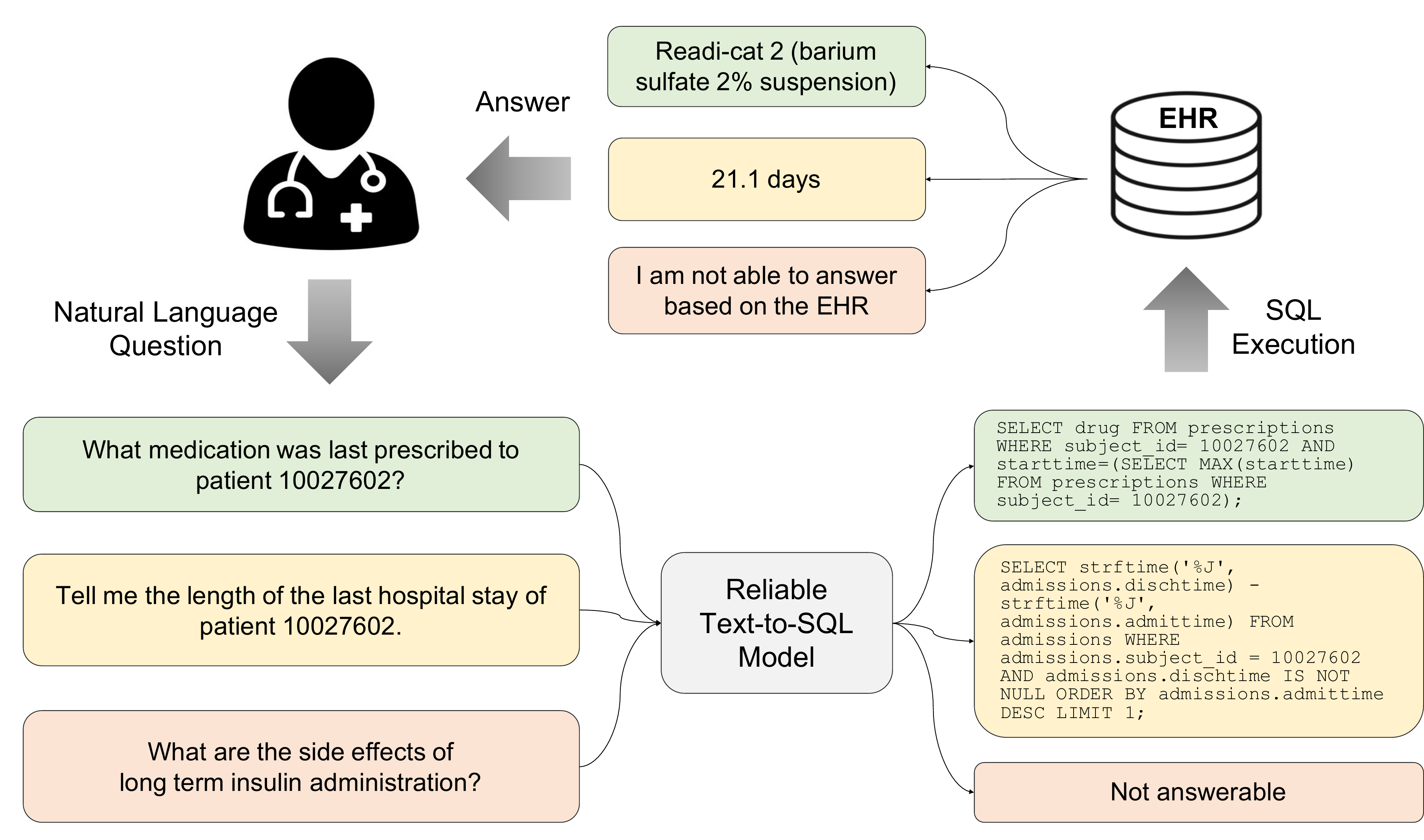}
\caption{Overview of reliable text-to-SQL modeling on EHRs. For any input questions, a reliable text-to-SQL model should accurately predict SQL queries for what it can and abstain from what it cannot, such as for intrinsically unanswerable questions or ones that are likely to be answered incorrectly by the model. Successfully developing such a model can serve as a valuable tool for healthcare professionals in hospitals, providing better accessibility to patient data and assisting in clinical decision-making.}
\label{fig:problem_formulation}
\end{figure*}

% It is the first dataset to compile a collection of questions that reflect the diverse needs of diverse healthcare professionals, based on a poll conducted at a university hospital. 
% This dataset covers two open-source EHR datasets, MIMIC-III~\cite{johnson2016mimic} and eICU~\cite{pollard2018eicu}, for annotated SQL queries.

Several datasets for question-answering on EHRs have been introduced, including MIMICSQL~\cite{wang2020text}, emrKBQA~\cite{raghavan-etal-2021-emrkbqa}, and EHRSQL~\cite{lee2022ehrsql}. EHRSQL, in particular, poses unique challenges. 
It is the first dataset to compile a collection of questions that reflect the diverse needs of healthcare professionals, including physicians, nurses, and hospital administrative staff.
It contains extensive use of time expressions and includes SQL queries of increased complexity, which better reflect the real needs of a hospital setting. 
The SQL queries are linked to two open-source EHR databases\footnote{SQL queries are database-dependent, meaning that even though a question attempts to retrieve the same information, the location of that information can vary across databases. For example, to list all drugs in MIMIC-III, you would use \texttt{SELECT drug FROM prescriptions}, whereas in eICU, it would be \texttt{SELECT drugname FROM medication}.},
MIMIC-III~\cite{johnson2016mimic} and eICU~\cite{pollard2018eicu}, retaining incompatible ones as unanswerable questions in the dataset (used to test a model's ability to abstain).
Starting from their collected real-world questions, this shared task presents more up-to-date changes to the text-to-SQL modeling (use of MIMIC-IV and new paraphrases for questions) and more challenging problem settings (new data splitting and additional unanswerable questions). The dataset for this shared task is publicly available at \url{https://github.com/glee4810/ehrsql-2024}. 
The shared task platform is hosted on Codabench at \url{https://www.codabench.org/competitions/1889/}.

In this paper, we present the EHRSQL 2024 shared task and its dataset in Sections 2 and 3, respectively. Section 4 introduces the evaluation metric and baseline model for the task. Section 5 describes the methods proposed by the participating teams and discusses interesting findings from the official results.

\begin{figure*}[t!]
\centering
\includegraphics[width=\textwidth]{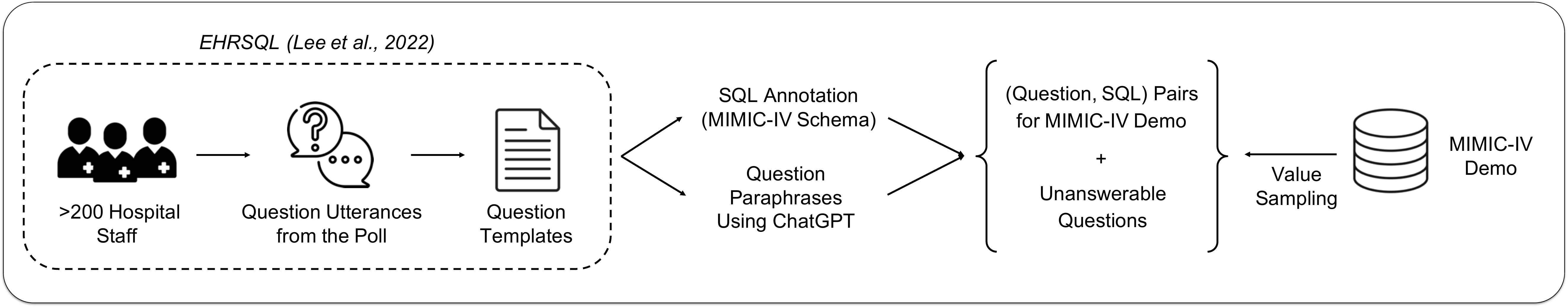}
\caption{Data construction process of the EHRSQL shared task.}
\label{fig:data_construction}
\end{figure*}

\section{Task - Reliable Text-to-SQL Modeling}

The goal of the EHRSQL 2024 shared task is to develop a reliable QA system for EHRs, specifically through text-to-SQL modeling.
Reliability is crucial for the deployment of AI systems, especially in safety-critical domains like hospitals, where incorrect predictions can have severe consequences.
The term reliability in question answering refers to the system's preference for abstention over providing an incorrect answer~\cite{whitehead2022reliable, chen2023adaptation, lee2024trustsql}.
In this shared task, we adopt the definition of \textit{reliable text-to-SQL} from TrustSQL~\cite{lee2024trustsql}, which expands the scope of reliability to include unanswerable questions.
A reliable text-to-SQL model should not only correctly generate SQL queries, providing utility, but also abstain from answering questions that are likely to be incorrect or unanswerable, thereby minimizing harm.
This objective contrasts with most other text-to-SQL tasks, where the primary focus is to maximize SQL generation performance for answerable questions only.
Further discussion on specific scenarios of measuring reliability for text-to-SQL is explained in Section~\ref{subsection:evaluation_metric}.

\section{Dataset Construction}
\label{sec:data_construction}

In this section, we outline the key steps to generate data for the shared task. The overall data construction pipeline is illustrated in Figure~\ref{fig:data_construction}. Each subsection provides a detailed explanation of each step.

\subsection{Question Templates from EHRSQL}

To construct the shared task data, we started from a pool of questions that reflect the real needs of diverse healthcare professionals in EHRSQL~\cite{lee2022ehrsql}. This dataset is derived from the results of a poll conducted with more than 200 professionals at a university hospital in South Korea. The collected questions are those that the professionals would ask an AI speaker if it could access and synthesize structured information stored in EHRs (i.e., records in tabular form). The authors then translated the raw question utterances and removed duplicates to distill them into question templates. This shared task leverages the question templates collected in EHRSQL to generate diverse and realistic question-SQL pairs.

\subsection{SQL Queries Linked to MIMIC-IV Demo}

Unlike the original EHRSQL dataset, whose SQL queries are based on value-shuffled MIMIC-III and eICU\footnote{This process was done to further de-identify the question-SQL pairs for public release. Please refer to more detailed reasons in the original paper.}, this shared task uses the demo version of MIMIC-IV\footnote{\url{https://physionet.org/content/mimic-iv-demo/2.2/}}~\cite{johnson2020mimic} to construct question-SQL pairs. 
The demo version contains records of 100 patients from the full MIMIC-IV database and has the same database schema as the full MIMIC-IV. It is openly available for anyone who is interested in using the dataset without special training\footnote{The full MIMIC-IV dataset requires researchers to complete the Collaborative Institutional Training Initiative (CITI) training before accessing the data.}, which allows the use of commercial APIs. Since the demo database schema is identical to that of the full database, the same query can be used to retrieve information from both the full and demo versions.

\subsection{New Question Paraphrases}

We found that the style and naturalness of paraphrases generated by current LLMs, like ChatGPT, surpass the paraphrases in EHRSQL, which are produced through both human and machine efforts.
To improve the quality of the paraphrases for each question template, we employed ChatGPT to generate new paraphrases that are more natural and conversational.
We then manually reviewed all new paraphrases to ensure they maintain the intended meaning of the original question templates.

\subsection{Challenging Unanswerable Questions}
\label{sec:unans}

A recent study revealed that unanswerable questions in the EHRSQL dataset can mostly be filtered out using a combination of N-gram and beam search score filtering~\cite{anonymous2024towards}. 
This is primarily because the unanswerable questions in EHRSQL were collected erroneously due to human errors during the polling process\footnote{The poll participants were initially provided with examples of inappropriate questions for the system, including those requiring external knowledge, ambiguous or qualitative statements, and questions about the reasons behind certain clinical decisions.}, resulting in limited diversity.
To increase the difficulty of the task, we combined the original unanswerable questions with those from the EHRSQL portion of TrustSQL~\cite{lee2024trustsql}, which contains adversarially created unanswerable questions, such as those referring to non-existent columns and requests that exceed SQL functionalities.

\subsection{New Data Split}

% In the original EHRSQL data, question templates are distributed across the training, validation, and test sets in an independent and identically distributed (IID) manner. 
% This ensures that all SQL structures are included in all splits, with each question template corresponding to a unique SQL structure.
% This setup indicates that if a model can accurately classify input questions into one of the question templates and generate the corresponding SQL queries, solving the problem becomes relatively easy. 

In real-world scenarios, text-to-SQL models can encounter questions that are answerable based on the EHR schema but have not been seen in the training set (i.e., unseen SQL with respect to the training set). 
This situation can lead to increased confusion for the model in distinguishing between answerable and unanswerable questions.
Unlike the original EHRSQL, where answerable questions were split in an identically distributed (IID) manner, we split the shared task data to include both seen and unseen question templates (or SQL structures) in the validation and test sets.
For unanswerable questions, the original unanswerable questions from EHRSQL were distributed across all splits (training, validation, and test), while new unanswerable questions were added exclusively to the validation and test sets to increase the task's difficulty.
Each of these splits has a 20\% proportion of unanswerable questions. 
Table~\ref{tab:data_statistic} shows the number of question templates and the size of each data split\footnote{Even if the MIMIC-IV demo includes only 100 patients, a wide variety of question templates can exist. Consider patient ID 100 and two question templates: `What is patient 100's gender?' and `What is patient 100's last blood pressure?' The data splitting in text-to-SQL for EHRs does not have to be done by patient, such as `What is patient 100's gender?' in the training set and `Tell me patient 200's sex?' in the validation set, because the task could become relatively easy. Instead, it might include `What is patient 100's gender?' in the training set and `What is patient 100's last blood pressure?' in the validation set. A more challenging and realistic goal of text-to-SQL is to assess how well the model can generate SQL queries for both question templates (or SQL structures) that it has seen and those it has not seen. In this example, we show four question samples with two question templates.}.
The training and validation sets were made available during the development phase (Jan 29, 2024 - Mar 26, 2024), and the test set was made available for the three-day test phase (Mar 26, 2024 - Mar 28, 2024).

\begin{table}[t!]
\centering
\Huge
\renewcommand{\arraystretch}{1.8}
\begin{adjustbox}{width=\linewidth,center}
\begin{tabular}{cccc}
\hline
& \multicolumn{2}{c}{Dev Phase} & Test Phase \\
\cmidrule(lr){2-3}
\cmidrule(lr){4-4}
& Train & Valid & Test \\
\hline
\makecell{Answerable \\ question template} & \makecell{100 \\ (100 seen)} & \makecell{134 \\ (100 seen + 34 unseen)} & \makecell{134 \\ (100 seen + 34 unseen)} \\
\hline
\makecell{Answerable \\ samples} & 4674 & 931 & 934 \\
\hline
\makecell{Unanswerable \\ samples} & 450 & 232 & 233 \\
\hline
Total samples & 5124 & 1163 & 1167 \\
\hline
\end{tabular}
\end{adjustbox}
\caption{Data statistics for the shared task. All text-to-SQL data used in the shared task is based on MIMIC-IV.}
\label{tab:data_statistic}
\end{table}

\begin{table*}[t!]
\centering
\small
\renewcommand{\arraystretch}{2.0}
% \begin{adjustbox}{width=\textwidth,center}
\begin{threeparttable}[t]
\begin{tabular}{ccccc}
\hline
& Team & Affiliation & Paper & Code \\
\hline
1 & LG AI Research \& KAIST & LG AI Research \& KAIST, South Korea & \citet{lg_ai} & \tnote{1} \\
2 & PromptMind & - & \citet{promptmind} & \tnote{2} \\
3 & ProbGate & KAIST, South Korea & \citet{probgate} & \tnote{3} \\
4 & KU-DMIS & Korea university, South Korea & \citet{ku_dmis} & \tnote{4} \\
5 & AIRI NLP & AIRI, Russia & \citet{airi} & \tnote{5} \\
6 & LTRC-IIITH & IIIT Hyderabad, India & \citet{ltrc} & \tnote{6} \\
7 & Saama Technologies & Saama Technologies, USA & \citet{saama} & \tnote{7} \\
8 & Project PRIMUS & SUST, Bangladesh & \citet{Project_PRIMUS} & \tnote{8} \\
\hline
\end{tabular}
\begin{tablenotes}
\item[1] \url{https://github.com/sylee0520/ehrsql-2024} (private)
\item[2] \url{https://github.com/satyakesav/ehrsql-clinicalnlp-2024} (private)
\item[3] \url{https://github.com/venzino-han/probgate_ehrsql}
\item[4] \url{https://github.com/Chanwhistle/EHRSQL_NACCL}
\item[5] \url{https://github.com/runnerup96/EHRSQL-text2sql-solution}
\item[6] \url{https://github.com/jr-john/ehrsql_2024} (private)
\item[7] \url{https://github.com/upjabir/ehrsql_2024}
\item[7] \url{https://github.com/joy-2019331037/nlpConference}
\end{tablenotes}
\end{threeparttable}
% \end{adjustbox}
\caption{Participating teams, affiliation, paper, and code.}
\label{tab:teams}
\end{table*}

\section{Evaluation}

\subsection{Evaluation Metric}
\label{subsection:evaluation_metric}

% We chose the evaluation metric that best aligns with the purpose of our shared task: to build reliable text-to-SQL models aimed at accurately predicting correct SQL queries and identifying unanswerable questions, while minimizing incorrect SQL predictions and the false detection of unanswerable questions as answerable.
We selected the evaluation metric that best aligns with our shared task’s purpose: building reliable text-to-SQL models. These models aim to accurately predict correct SQL queries for answerable questions, identify unanswerable questions, and minimize both incorrect SQL predictions for answerable questions and the generation of SQL queries for unanswerable questions, without abstaining.
More concretely, we adopt the Reliability Score (RS) for reliable text-to-SQL~\cite{lee2024trustsql}, formally written as follows:

\begin{equation}
\small
RS(c)(x) = 
    \begin{cases}
      1 & \text{if $x \in \mathcal{Q}_{ans}$; $g(x) = 1$; $Acc(x) = 1$, }\\
      0 & \text{if $x \in \mathcal{Q}_{ans}$; $g(x) = 0$,}\\   
      -c & \text{if $x \in \mathcal{Q}_{ans}$; $g(x) = 1$; $Acc(x) = 0$,}\\
      -c & \text{if $x \in \mathcal{Q}_{una}$; $g(x) = 1$,}\\
      1 & \text{if $x \in \mathcal{Q}_{una}$; $g(x) = 0$,}\\
    \end{cases}
\label{eq:reliability-metric}
\end{equation}

\noindent where $\mathcal{Q}_{ans}$ and $\mathcal{Q}_{una}$ denote answerable and unanswerable questions, respectively. g(x) = 1 implies that the model selects its SQL generation as the final answer, whereas g(x) = 0 implies that the model abstains.
Acc(x) measures the accuracy of the generated SQL using execution accuracy, which is 1 if the answers returned by both the ground-truth and predicted SQL queries match, and 0 otherwise.

The RS has five different cases for assigning the score:

\begin{itemize} % [itemsep=-0.5pt, topsep=-0.5pt]
  \item A score of $1$ is assigned if SQL is correctly generated by the model for answerable questions.
  \item A score of $0$ is assigned if the model abstains from generating SQL for answerable questions.
  \item A score of $-c$ is assigned if the model predicts incorrect SQL for answerable questions.
  \item A score of $-c$ is assigned if the model attempts to predict SQL for unanswerable questions.
  \item A score of $1$ is assigned if the model accurately detects unanswerable questions by abstaining.
\end{itemize}

The overall RS is calculated by taking the average of sample-level scores, represented as percentages.
The penalty, c, is chosen depending on the reliability requirements of the model.
A penalty of 0, RS(0), means no punishment for incorrect predictions, a penalty of 10; RS(10), represents a moderately rigorous scenario, and a penalty of N; RS(N), where N refers to the evaluation data size, is the most rigorous scenario in which even a single mistake outweighs all correct predictions and abstentions.
The maximum possible RS is $100\%$, and the minimum possible scores vary depending on the penalties: $0$ for $c=0$; $-1000\%$ for $c=10$; $-100N\%$ for $c=N$.
The main evaluation metric for the shared task is RS(10), where every ten accurate predictions weigh the same as one incorrect prediction.

\subsection{Code Verification and Fact Sheet}

The participants shared their code and the fact sheet following the instructions provided in Appendix~\ref{code_submission_instruction}. 
The purpose of the fact sheet was to collect a brief summary of the participants' methods, including any use of pre-trained models or external data. 
For code verification, participants had the option to submit their code either via email or through GitHub repositories.
These repositories could be public or private, as long as access was granted to the task organizers. 
Upon receiving the submissions, we conducted a careful review to ensure that the provided code and the methods described in the fact sheets were consistent.

\begin{table*}[t!]
\centering
% \normalsize
\renewcommand{\arraystretch}{2.0}
\begin{adjustbox}{width=\textwidth,center}
\begin{tabular}{ccccccccc}
\hline
& Team & RS(0) & RS(10) & RS(N) & Modeling Type & Ensemble & Fine-tuned & Model Used \\
\hline
1 & LG AI Research \& KAIST & 88.17 &  81.32 & -711.83 & Unified & No & Yes  & ChatGPT \\ 
2 & PromptMind & 82.6 & 74.89 & -817.4 & Unified & Yes & Yes & GPT-4, ChatGPT, Claude Opus \\
3 & ProbGate & 81.92 & 74.21 & -818.08 & Unified & No & Yes & ChatGPT \\
4 & KU-DMIS & 72.07 & 59.21 & -1427.93 & Unified & No & Yes & ChatGPT \\
5 & AIRI NLP & 68.89 & 44.04 & -2831.11 & Pipeline & No & Yes & T5-3B, Logistic Regression \\
6 & LTRC-IIITH & 66.84 & 43.7 & -2633.16 & Pipeline & No & Yes & SQLCoder-7b-2 \\
7 & Saama Technologies & 53.21 & 36.08 & -1946.79 & Pipeline & Yes & Yes & Decision Trees, CodeLlama-7b, ChatGPT \\
8 & Project PRIMUS & 14.14 & -713.37 & -84.9K & Unified & No & No & SQLCoder-7b-2 \\
\hline
- & \textsc{Abstain-All} & 20.0 & 20.0 & 20.0 & No & No & No & - \\
\hline
\end{tabular}
\end{adjustbox}
\caption{Official result. \textsc{Abstain-All} is the baseline for the shared task, explained in Section~\ref{sec:baseline}. `Ensemble' denotes the use of any ensemble methods in either unanswerable question detection or SQL generation with abstention. `Fine-tuned' indicates whether any pre-trained models were further trained for SQL generation or abstention purposes. `Pipeline-based' means the use of multiple methods in a sequence, such as a pipeline that consists of an answerability detector, an SQL generator, and subsequently an SQL error detector.}
\label{tab:methods}
\end{table*}

\subsection{Baseline Model}
\label{sec:baseline}

For the baseline, we employ the simplest method, denoted as \textsc{Abstain-All}, which abstains from answering all questions. 
When evaluating in the RS, regardless of the penalty, abstaining from all questions results in an overall score of 20\%. 
This score is earned by correctly abstaining from answering unanswerable questions. 
This is not a trivial score, particularly as the penalty for incorrect predictions increases, which can severely harm the overall score.

\section{Official Results}

\subsection{Participating Teams}

The EHRSQL shared task attracted over 100 participants from both academia and industry.
Of these, 8 teams submitted their code and fact sheet.
Table~\ref{tab:teams} lists the participating teams, their affiliations, the code submission status, and their working papers.

\subsection{Methods and Results}

Table~\ref{tab:methods} presents the official results for each team, along with short descriptions of their methods. All the proposed methods are categorized into two types: unified approaches and pipeline-based approaches (see `Modeling Type' in Table~\ref{tab:methods}). The unified approach leverages LLMs to perform both SQL generation and abstention, while the pipeline-based approach involves building a series of specialized, smaller models to ensure reliability as one system (e.g., detection of unanswerable questions, followed by SQL generation and error detection). The overall observations are as follows: 1) Even when utilizing the same backbone LLM (e.g., ChatGPT), the results vary depending on how it is used; 2) Methods that fall under the unified approach tend to outperform those in the pipeline-based approaches. However, leveraging LLMs in a pipeline-based manner has not been fully explored; 3) Teams with smaller discrepancies between the RS with different penalties (e.g., the gap between RS(0) and RS(10)) tend to rank higher, indicating that minimizing incorrect SQL predictions through effective abstention mechanisms is crucial for higher performance in this task; 4) Teams that fine-tuned LLMs on the training data, whether they used general-purpose (e.g., ChatGPT) or code-specialized models (e.g., CodeLLama), performed well, highlighting the importance of domain-specific fine-tuning in this task; 5) Test-time pseudo-label fine-tuning seems quite effective in fixed benchmarking scenarios, which might not be directly applicable in real-time deployment. More details about each method by the two different modeling types are discussed in the following paragraphs.

\paragraph{Unified approach.} Five teams utilized methods under the unified approach. The LG AI Research \& KAIST team achieved the best results, scoring 81.32 in RS(10) by using self-training LLMs~\cite{amini2022self, yuan2024self} with pseudo-labeling for unanswerable questions. The PromptMind team implemented an ensemble of LLMs, including fine-tuned ChatGPT, GPT-4, and Claude Opus. When generating SQL, they employed two retrievers (one for the general domain and another for the medical domain) to retrieve similar question-SQL pairs from the training set. They selected SQL generation as the final prediction only if all three models unanimously agreed; otherwise, the SQL prediction was abstained. The ProbGate team employed fine-tuned ChatGPT with log-probability thresholding and error handling for abstention, where the threshold was set heuristically based on the ratio of unanswerable questions in the validation set. The KU-DMIS team took a two-step procedure. First, they generated pseudo question-SQL pairs to align their models to the test set distribution using the training set and question templates from the original EHRSQL data. Then, they fine-tuned ChatGPT on the newly generated dataset. Abstention was achieved by sampling multiple SQL predictions for each input question and abstaining if the outputs were not consistent. The Project PRIMUS team used SQLCoder-7b-2 for direct generation of SQL and abstention labels (null) through in-context learning.

\paragraph{Pipeline-based approach.} Three teams adopted the pipeline-based approach. The AIRI NLP team constructed a pipeline where they initially used logistic regression to detect unanswerable questions, then generated SQL with a fine-tuned T5-3B~\cite{raffel2020exploring}, and finally checked the executability of the generated SQL for final abstention. The LTRC-IIITH team used two different SQLCoder-7b-2 models, one for detecting unanswerable questions followed by the other for SQL generation. For final abstention, they utilized the log-probabilities from the SQL generator to detect potential errors in SQL generation, followed by an executability check of the SQL. The Saama Technologies team began with an ensemble of unanswerable question detectors, including multinomial naive Bayes, SGD classifier, CatBoost~\cite{prokhorenkova2018catboost}, and CodeLlama-7b~\cite{roziere2023code}. They then generated SQL using CodeLlama-7b and finally used a ChatGPT-based answer selector for final abstention.

\section{Conclusion}

With the increasing volume of data stored in EHRs and the impressive advances in LLMs, the EHRSQL 2024 shared task provides participants with an opportunity to develop and test their creative methods for building reliable QA systems on EHRs using text-to-SQL modeling. The dataset for this shared task presents unique challenges, including questions that extensively use time expressions and the increased complexity of SQL queries, reflecting the varied needs of different hospital professionals. It also includes challenging unanswerable questions that should be avoided to enhance the system’s reliability. These aspects distinguish this shared task from most other text-to-SQL challenges, as it requires not only generating correct SQL queries that reflect the actual needs (providing utility) but also abstaining from answering questions that are likely to be answered incorrectly or are unanswerable (minimizing harm), which is especially crucial in the real deployment of the system in clinical settings.

The shared task attracted over 100 participants from academia and industry, with 8 teams ultimately submitting their code and fact sheets. As a novel task at the intersection of NLP and clinical domains, it triggered a variety of newly proposed methods. These methods included self-training LLMs through pseudo-labeling unanswerable questions, ensembling different LLMs with an abstention option, generating synthetic question-SQL pairs to handle distribution shifts from training to test sets, leveraging log-probabilities for abstention, and deploying pipeline-based approaches with specialized models for correct SQL generation and abstention. Most teams achieved positive scores (more utility than harm) in the moderately rigorous setting (i.e., RS(10), where one mistake weighs as much as ten correct predictions), which was the main evaluation metric.

However, none of the proposed models have yet achieved positive scores in RS(N) (where even one incorrect prediction outweighs the rest of the predictions being correct), indicating that they are still far from being completely reliable. We hope that this shared task, emphasizing reliability, will encourage further research into building QA systems that can truly serve as valuable tools for healthcare professionals, improving clinical decision-making, facilitating clinical research, and enhancing patient care quality. Future research directions include developing models that can achieve positive scores in RS(N) and expanding reliable question answering for EHRs to multimodal settings by incorporating clinical notes, X-ray images, and ECG signals.

\section*{Limitations}

This shared task does not represent all types of answerable and unanswerable questions encountered in hospital settings. Additionally, this shared task employs MIMIC-IV as the EHR database, which is not a universally accepted EHR schema, and the databases are preprocessed for the QA task by eliminating duplicate values across different tables to reduce ambiguity. Lastly, further experiments are necessary for newly proposed LLMs, since most methods, including text-to-SQL generation and abstention, depend heavily on the underlying LLMs.

\section*{Acknowledgments}
We would like to thank Asma Ben Abacha from Microsoft Health AI and the ClinicalNLP organizers for their feedback and support for the EHRSQL 2024 shared task. We also thank all the participating teams who contributed to the success of this shared task by proposing their interesting methods and showing strong engagement.
This work was supported by the Institute for Information \& communications Technology Promotion(IITP) grant (No.2019-0-00075) and the National Research Foundation of Korea (NRF) grant (NRF-2020H1D3A2A03100945) funded by the Korea government (MSIT).

\bibliography{custom}

\newpage
\appendix

\onecolumn
\section{Code Submission and Fact Sheet Template}
\label{sec:appendix}

\label{code_submission_instruction}
\input{appendix/code_submission_instruction}

\end{document}

%% file: appendix/code_submission_instruction.tex
\Large
\textbf{Fact Sheet for EHRSQL-2024 Shared Task:}\\

\large
\textbf{I. Team name}

\begin{itemize} % [itemsep=-0.5pt, topsep=-0.5pt] 
  \item Username on Codalab:
  \item Team leader affiliation:
  \item Team leader email:
  \item Name of other team members (and affiliation):
  \item Team website URL (if any): \newline
\end{itemize}

\textbf{II. Contribution}
\begin{itemize} % [itemsep=-0.5pt, topsep=-0.5pt \ding{109}]
  \item \textbf{Title of the contribution}
    \begin{itemize}
    \item Provide a brief summary of the method and contributions.
    \end{itemize}
  \item \textbf{Representative image / workflow diagram of the method}
    \begin{itemize}
    \item An image (or several images) to support method description to better understand the approach and model pipeline. You can refer to these images in the method description part.
    \end{itemize}
  \item \textbf{Detailed method description}
    \begin{itemize}
    \item Provide a technical and detailed description of the method and contributions. The explanations must be self-contained and one must be able to reproduce the approach by reading this section.
    \end{itemize}
  \item \textbf{Shared task results}
    \begin{itemize}
    \item $RS_{0}$:
    \item $RS_{5}$:
    \item $RS_{10}$:
    \item $RS_{N}$:
    \end{itemize}
  \item \textbf{Final Remarks}
    \begin{itemize}
    \item Please identify the pros and cons (if any) of the proposed approach. \newline
    \end{itemize}
\end{itemize}

\textbf{III. Additional method details}

\begin{itemize} % [itemsep=-0.5pt, topsep=-0.5pt] 
  \item Did you use any pre-trained model?
  \item Did you use external data?
  \item Did you perform any data augmentation?
  \item At the test phase, did you use the provided validation set as part of your training set?
  \item Did you use any regularization strategies/terms?
  \item Did you use handcrafted features?
  \item Did you use any domain adaptation strategy?
  \newline
\end{itemize}

\textbf{IV. Code Repository}

\begin{itemize} % [itemsep=-0.5pt, topsep=-0.5pt] 
  \item Link to a code repository with complete and detailed instructions so that the results obtained on Codabench can be reproduced.
  \item If private repo, share the repo with \texttt{glee4810}
\end{itemize}

%% file: acl_latex.bbl
\begin{thebibliography}{30}
\providecommand{\natexlab}[1]{#1}

\bibitem[{Amini et~al.(2022)Amini, Feofanov, Pauletto, Devijver, and Maximov}]{amini2022self}
Massih-Reza Amini, Vasilii Feofanov, Loic Pauletto, Emilie Devijver, and Yury Maximov. 2022.
\newblock Self-training: A survey.
\newblock \emph{arXiv preprint arXiv:2202.12040}.

\bibitem[{Chang and Fosler-Lussier(2023)}]{chang2023prompt}
Shuaichen Chang and Eric Fosler-Lussier. 2023.
\newblock How to prompt llms for text-to-sql: A study in zero-shot, single-domain, and cross-domain settings.
\newblock In \emph{NeurIPS 2023 Second Table Representation Learning Workshop}.

\bibitem[{Chen et~al.(2023)Chen, Yoon, Ebrahimi, Arik, Pfister, and Jha}]{chen2023adaptation}
Jiefeng Chen, Jinsung Yoon, Sayna Ebrahimi, Sercan Arik, Tomas Pfister, and Somesh Jha. 2023.
\newblock Adaptation with self-evaluation to improve selective prediction in llms.
\newblock In \emph{Findings of the Association for Computational Linguistics: EMNLP 2023}, pages 5190--5213.

\bibitem[{Gao et~al.(2023)Gao, Wang, Li, Sun, Qian, Ding, and Zhou}]{gao2023text}
Dawei Gao, Haibin Wang, Yaliang Li, Xiuyu Sun, Yichen Qian, Bolin Ding, and Jingren Zhou. 2023.
\newblock Text-to-sql empowered by large language models: A benchmark evaluation.
\newblock \emph{arXiv preprint arXiv:2308.15363}.

\bibitem[{Gundabathula and Kolar(2024)}]{promptmind}
Satya Gundabathula and Sriram Kolar. 2024.
\newblock Promptmind team at ehrsql-2024: Improving reliability of sql generation using ensemble llms.
\newblock In \emph{Proceedings of the 6th Clinical Natural Language Processing Workshop}, Mexico City, Mexico. Association for Computational Linguistics.

\bibitem[{Jabir et~al.(2024)Jabir, Kanakarajan, and Sankarasubbu}]{saama}
Mohammed Jabir, Kamal Kanakarajan, and Malaikannan Sankarasubbu. 2024.
\newblock Saama technologies at ehrsql 2024: Sql generation through classification answer selector by llm.
\newblock In \emph{Proceedings of the 6th Clinical Natural Language Processing Workshop}, Mexico City, Mexico. Association for Computational Linguistics.

\bibitem[{Jo et~al.(2024)Jo, Lee, Seo, Hwang, and Lee}]{lg_ai}
Yongrae Jo, Seongyun Lee, Minju Seo, Sung~Ju Hwang, and Moontae Lee. 2024.
\newblock Lg ai research \& kaist at ehrsql 2024: Self-training large language models with pseudo-labeled unanswerable questions for a reliable text-to-sql system on ehrs.
\newblock In \emph{Proceedings of the 6th Clinical Natural Language Processing Workshop}, Mexico City, Mexico. Association for Computational Linguistics.

\bibitem[{Johnson et~al.(2020)Johnson, Bulgarelli, Pollard, Horng, Celi, and Mark}]{johnson2020mimic}
Alistair Johnson, Lucas Bulgarelli, Tom Pollard, Steven Horng, Leo~Anthony Celi, and Roger Mark. 2020.
\newblock Mimic-iv.
\newblock \emph{PhysioNet. Available online at: https://physionet. org/content/mimiciv/1.0/(accessed August 23, 2021)}, pages 49--55.

\bibitem[{Johnson et~al.(2016)Johnson, Pollard, Shen, Lehman, Feng, Ghassemi, Moody, Szolovits, Anthony~Celi, and Mark}]{johnson2016mimic}
Alistair~EW Johnson, Tom~J Pollard, Lu~Shen, Li-wei~H Lehman, Mengling Feng, Mohammad Ghassemi, Benjamin Moody, Peter Szolovits, Leo Anthony~Celi, and Roger~G Mark. 2016.
\newblock Mimic-iii, a freely accessible critical care database.
\newblock \emph{Scientific data}, 3(1):1--9.

\bibitem[{Joy et~al.(2024)Joy, Ahmed, Saha, Ahmed, Das, and Bhowmik}]{Project_PRIMUS}
Sourav~Bhowmik Joy, Rohan Ahmed, Argha~Pratim Saha, Minhaj Ahmed, Utsho Das, and Partha~Sarothi Bhowmik. 2024.
\newblock Project primus at ehrsql 2024: Text-to-sql generation using large language model for ehr analysis.
\newblock In \emph{Proceedings of the 6th Clinical Natural Language Processing Workshop}, Mexico City, Mexico. Association for Computational Linguistics.

\bibitem[{Kim et~al.(2024{\natexlab{a}})Kim, Kim, Lee, Lee, Jang, Lee, Kim, and Kang}]{ku_dmis}
Chanhwi Kim, Hajung Kim, Hoonick Lee, Jiwoo Lee, Kyochul Jang, Kyungjae Lee, Gangwoo Kim, and Jaewoo Kang. 2024{\natexlab{a}}.
\newblock Ku-dmis: Generating sql query via question templatization in ehr.
\newblock In \emph{Proceedings of the 6th Clinical Natural Language Processing Workshop}, Mexico City, Mexico. Association for Computational Linguistics.

\bibitem[{Kim et~al.(2024{\natexlab{b}})Kim, Han, and Kim}]{probgate}
Sangryul Kim, Donghee Han, and Sehyun Kim. 2024{\natexlab{b}}.
\newblock Probgate at ehrsql 2024: Enhancing sql query generation accuracy through probabilistic threshold filtering and error handling.
\newblock In \emph{Proceedings of the 6th Clinical Natural Language Processing Workshop}, Mexico City, Mexico. Association for Computational Linguistics.

\bibitem[{Lee et~al.(2024)Lee, Chay, Cho, and Choi}]{lee2024trustsql}
Gyubok Lee, Woosog Chay, Seonhee Cho, and Edward Choi. 2024.
\newblock Trustsql: A reliability benchmark for text-to-sql models with diverse unanswerable questions.
\newblock \emph{arXiv preprint arXiv:2403.15879}.

\bibitem[{Lee et~al.(2022)Lee, Hwang, Bae, Kwon, Shin, Yang, Seo, Kim, and Choi}]{lee2022ehrsql}
Gyubok Lee, Hyeonji Hwang, Seongsu Bae, Yeonsu Kwon, Woncheol Shin, Seongjun Yang, Minjoon Seo, Jong-Yeup Kim, and Edward Choi. 2022.
\newblock Ehrsql: A practical text-to-sql benchmark for electronic health records.
\newblock \emph{Advances in Neural Information Processing Systems}, 35:15589--15601.

\bibitem[{Mullins et~al.(2020)Mullins, O’Donnell, Mousa, Rankin, Ben-Meir, Boyd-Skinner, and Skouteris}]{mullins2020health}
Alexandra Mullins, Renee O’Donnell, Mariam Mousa, David Rankin, Michael Ben-Meir, Christopher Boyd-Skinner, and Helen Skouteris. 2020.
\newblock Health outcomes and healthcare efficiencies associated with the use of electronic health records in hospital emergency departments: a systematic review.
\newblock \emph{Journal of Medical Systems}, 44(12):200.

\bibitem[{Pollard et~al.(2018)Pollard, Johnson, Raffa, Celi, Mark, and Badawi}]{pollard2018eicu}
Tom~J Pollard, Alistair~EW Johnson, Jesse~D Raffa, Leo~A Celi, Roger~G Mark, and Omar Badawi. 2018.
\newblock The eicu collaborative research database, a freely available multi-center database for critical care research.
\newblock \emph{Scientific data}, 5(1):1--13.

\bibitem[{Pourreza and Rafiei(2024)}]{pourreza2024din}
Mohammadreza Pourreza and Davood Rafiei. 2024.
\newblock Din-sql: Decomposed in-context learning of text-to-sql with self-correction.
\newblock \emph{Advances in Neural Information Processing Systems}, 36.

\bibitem[{Prokhorenkova et~al.(2018)Prokhorenkova, Gusev, Vorobev, Dorogush, and Gulin}]{prokhorenkova2018catboost}
Liudmila Prokhorenkova, Gleb Gusev, Aleksandr Vorobev, Anna~Veronika Dorogush, and Andrey Gulin. 2018.
\newblock Catboost: unbiased boosting with categorical features.
\newblock \emph{Advances in neural information processing systems}, 31.

\bibitem[{Raffel et~al.(2020)Raffel, Shazeer, Roberts, Lee, Narang, Matena, Zhou, Li, and Liu}]{raffel2020exploring}
Colin Raffel, Noam Shazeer, Adam Roberts, Katherine Lee, Sharan Narang, Michael Matena, Yanqi Zhou, Wei Li, and Peter~J Liu. 2020.
\newblock Exploring the limits of transfer learning with a unified text-to-text transformer.
\newblock \emph{Journal of machine learning research}, 21(140):1--67.

\bibitem[{Raghavan et~al.(2021)Raghavan, Liang, Mahajan, Chandra, and Szolovits}]{raghavan-etal-2021-emrkbqa}
Preethi Raghavan, Jennifer~J Liang, Diwakar Mahajan, Rachita Chandra, and Peter Szolovits. 2021.
\newblock \href {https://doi.org/10.18653/v1/2021.bionlp-1.7} {emr{KBQA}: A clinical knowledge-base question answering dataset}.
\newblock In \emph{Proceedings of the 20th Workshop on Biomedical Language Processing}, pages 64--73, Online. Association for Computational Linguistics.

\bibitem[{Roziere et~al.(2023)Roziere, Gehring, Gloeckle, Sootla, Gat, Tan, Adi, Liu, Remez, Rapin et~al.}]{roziere2023code}
Baptiste Roziere, Jonas Gehring, Fabian Gloeckle, Sten Sootla, Itai Gat, Xiaoqing~Ellen Tan, Yossi Adi, Jingyu Liu, Tal Remez, J{\'e}r{\'e}my Rapin, et~al. 2023.
\newblock Code llama: Open foundation models for code.
\newblock \emph{arXiv preprint arXiv:2308.12950}.

\bibitem[{Scholak et~al.(2021)Scholak, Schucher, and Bahdanau}]{scholak2021picard}
Torsten Scholak, Nathan Schucher, and Dzmitry Bahdanau. 2021.
\newblock Picard: Parsing incrementally for constrained auto-regressive decoding from language models.
\newblock In \emph{Proceedings of the 2021 Conference on Empirical Methods in Natural Language Processing}, pages 9895--9901.

\bibitem[{Somov et~al.(2024)Somov, Tutubalina, and Dontsov}]{airi}
Oleg Somov, Elena Tutubalina, and Alexei Dontsov. 2024.
\newblock Airi nlp team at ehrsql 2024: T5 and logistic regression to the rescue.
\newblock In \emph{Proceedings of the 6th Clinical Natural Language Processing Workshop}, Mexico City, Mexico. Association for Computational Linguistics.

\bibitem[{Thomas et~al.(2024)Thomas, Mishra, Sharma, and Krishnamurthy}]{ltrc}
Jerrin~John Thomas, Pruthwik Mishra, Dipti Sharma, and Parameswari Krishnamurthy. 2024.
\newblock Ltrc-iiith at ehrsql 2024: Enhancing reliability of text-to-sql systems through abstention and confidence thresholding.
\newblock In \emph{Proceedings of the 6th Clinical Natural Language Processing Workshop}, Mexico City, Mexico. Association for Computational Linguistics.

\bibitem[{Upadhyay and Hu(2022)}]{upadhyay2022qualitative}
Soumya Upadhyay and Han-fen Hu. 2022.
\newblock A qualitative analysis of the impact of electronic health records (ehr) on healthcare quality and safety: Clinicians’ lived experiences.
\newblock \emph{Health Services Insights}, 15:11786329211070722.

\bibitem[{Uslu et~al.(2021)Uslu, Stausberg et~al.}]{uslu2021value}
Aykut Uslu, J{\"u}rgen Stausberg, et~al. 2021.
\newblock Value of the electronic medical record for hospital care: update from the literature.
\newblock \emph{Journal of medical Internet research}, 23(12):e26323.

\bibitem[{Wang et~al.(2020)Wang, Shi, and Reddy}]{wang2020text}
Ping Wang, Tian Shi, and Chandan~K Reddy. 2020.
\newblock Text-to-sql generation for question answering on electronic medical records.
\newblock In \emph{Proceedings of The Web Conference 2020}, pages 350--361.

\bibitem[{Whitehead et~al.(2022)Whitehead, Petryk, Shakib, Gonzalez, Darrell, Rohrbach, and Rohrbach}]{whitehead2022reliable}
Spencer Whitehead, Suzanne Petryk, Vedaad Shakib, Joseph Gonzalez, Trevor Darrell, Anna Rohrbach, and Marcus Rohrbach. 2022.
\newblock Reliable visual question answering: Abstain rather than answer incorrectly.
\newblock In \emph{European Conference on Computer Vision}, pages 148--166. Springer.

\bibitem[{Yang et~al.(2024)Yang, Kim, Kim, Lee, Yun, and Choi}]{anonymous2024towards}
Yongjin Yang, Sihyeon Kim, SangMook Kim, Gyubok Lee, Se-Young Yun, and Edward Choi. 2024.
\newblock \href {https://openreview.net/forum?id=Iv1h93UKYL} {Towards unbiased evaluation of detecting unanswerable questions in {EHRSQL}}.
\newblock In \emph{ICLR 2024 Workshop on Navigating and Addressing Data Problems for Foundation Models}.

\bibitem[{Yuan et~al.(2024)Yuan, Pang, Cho, Sukhbaatar, Xu, and Weston}]{yuan2024self}
Weizhe Yuan, Richard~Yuanzhe Pang, Kyunghyun Cho, Sainbayar Sukhbaatar, Jing Xu, and Jason Weston. 2024.
\newblock Self-rewarding language models.
\newblock \emph{arXiv preprint arXiv:2401.10020}.

\end{thebibliography}
